\documentclass[11pt,letterpaper]{article}
\usepackage{emnlp2016}
\emnlpfinalcopy
\usepackage{times}
\usepackage{latexsym}
\usepackage{url}
\usepackage{multirow}
\usepackage{graphicx}
\def\emnlppaperid{402}

\newcommand{\ignore}[1]{}

\title{Learning to Translate for Multilingual Question Answering}

\author{
Ferhan Ture\\
  Comcast Labs\Thanks{~This work was completed while author was an employee of Raytheon BBN Technologies.} \\
  1110 Vermont Ave NW Ste 600 \\
  Washington, DC, 20005 USA \\
  {\tt ferhan\_ture@cable.comcast.com} \\\And
  Elizabeth Boschee \\
  Raytheon BBN Technologies \\
  10 Moulton St \\
  Cambridge, MA, 02138 USA \\
  {\tt eboschee@bbn.com} \\
  }
  
\date{}

\begin{document}

%
%
%

\maketitle

\begin{abstract}

In multilingual question answering, either 
the question needs to be translated into the document language, or vice versa.
In addition to \emph{direction}, there are multiple \emph{methods} to perform the translation, four of which we 
explore in this paper:\ word-based, 10-best, context-based, and grammar-based. We 
build a feature for each combination of translation \emph{direction} and \emph{method}, 
and train a model that learns optimal feature weights.
On a large forum dataset consisting of posts in English, Arabic, and Chinese, 
our novel \emph{learn-to-translate} approach was more effective than a strong
baseline ($p<0.05$):\ translating all text into English, then training 
a classifier based only on English (original or translated) text.
\end{abstract}

\section{Introduction}

\emph{Question answering} (QA) is a specific form of the information retrieval (IR) task, where the 
goal is to find relevant well-formed answers to a posed question. Most QA pipelines consist of three main stages: (a) preprocessing the 
question and collection, (b) retrieval of candidate answers in the collection, and (c) ranking answers with respect to their relevance to 
the question and return the top $N$ answers. The types of 
questions can range from factoid (e.g., ``What is the capital of France?") to causal (e.g., ``Why are trees green?"), and opinion questions 
(e.g., ``Should USA lower the drinking age?"). 
The most common approach to \emph{multilingual QA} (MLQA) has been to translate all content into its most probable 
English translation via machine translation (MT) systems. This strong baseline, which we refer to as \emph{one-best MT} ({\tt 1MT}), has 
been successful in prior work~\cite{Hartrumpf:2009aa,Lin:2010aa,Shima:2010aa}. \ignore{\cite{Adafre:2009aa,Hartrumpf:2009aa}}
However, recent advances in cross-lingual IR (CLIR) show that one can do better by representing the translation space 
as a probability distribution~\cite{Ture:2014aa}. In addition, MT systems perform substantially worse with user-generated text, such as web 
forums~\cite{Wees:2015aa}, which provide extra motivation to consider alternative translation approaches for higher recall.
To our knowledge, it has yet to be shown whether these recent advancements in CLIR transfer to MLQA.


We introduce a novel answer ranking approach for MLQA (i.e., \emph{Learning to Translate} or {\tt L2T}), a model that 
learns the optimal translation of question and/or candidate answer, based on how well it discriminates between good and bad answers.
We achieve this by introducing a set of features that encapsulate lexical and semantic similarities between a question 
and a candidate answer through various translation strategies (Section~\ref{sec:repr}). The model then learns feature weights for 
each combination of translation \emph{direction} and \emph{method}, through a discriminative training process (Section~\ref{sec:feat}). 
\ignore{In addition to our novel features, we also experimented with various data selection strategies to optimize model training (Section~\ref{sec:data}).}
Once a model is trained, it can be used for MLQA, by sorting each candidate answer in the collection by 
model score. Instead of learning a single model to score candidate answers in any language, it might be meaningful to train a 
separate model that can learn to discriminate between good and bad answers in each language. This can let each model 
learn \emph{feature weights custom to the language}, therefore allowing a more fine-grained ranking (Section~\ref{sec:oneclassperlang}). 
We call this alternative approach \emph{Learning to Custom Translate} ({\tt L2CT}).


Experiments on the DARPA Broad Operational Language Technologies (BOLT) IR task\footnote{{\scriptsize \url{
http://www.darpa.mil/Our_Work/I2O/Programs}}} 
confirm that {\tt L2T} yields statistically significant improvements over a strong baseline ($p<0.05$),
in three out of four experiments. 
{\tt L2CT} outperformed the baseline as well, but was not more effective than {\tt L2T}. 
\ignore{We discuss 
other potential advantages of using language-specific classifiers.}


\section{Related Work}\label{sec:related}
For the last decade or so, research in QA has mostly been driven by annual evaluation campaigns like TREC,\footnote{{\scriptsize \url{http://trec.nist.gov}}}
 CLEF,\footnote{{\scriptsize \url{http://www.clef-initiative.eu}}} and NTCIR.\footnote{{\scriptsize \url{http://research.nii.ac.jp/ntcir/index.html}}} 
Most earlier work relied on either rule-based approaches where a set of rules were manually crafted for each type of question, or 
IR-like approaches where each pair of question and candidate answer was scored using retrieval functions (e.g., 
BM25~\cite{Robertson:2004}). On the other hand, training a classifier for ranking candidate answers allows the exploitation of 
various features extracted from the question, candidate answer, and surrounding context~\cite{Madnani:2007aa,Zhang:2007aa}. 
In fact, an explicit comparison at 2007 TREC confirmed the superiority of machine learning-based (ML-based) approaches (F-measure 35.9\%
vs 38.7\%)~\cite{Zhang:2007aa}. Learning-to-rank approaches have also been applied to QA successfully~\cite{Agarwal:2012aa}.

Previous ML-based approaches have introduced useful features from many aspects of natural language, 
including lexical~\cite{Brill:2001aa,Attardi:2001aa}, syntactic~\cite{Alfonseca:2001aa,Katz:2005aa}, 
semantic~\cite{Cui:2005aa,Katz:2005aa,Alfonseca:2001aa,Hovy:2001aa}, and discourse features, such as coreference 
resolution~\cite{Morton:1999aa}, or identifying temporal/spatial references~\cite{Saquete:2005aa,Harabagiu:2005aa},
which are especially useful for ``why'' and ``how'' questions~\cite{Kolomiyets:2011aa}.
Additionally, semantic role labeling and dependency trees are other forms of semantic analysis used widely in NLP
applications~\cite{Shen:2007aa,Cui:2005aa}.

When dealing with multilingual collections, most prior approaches translate all text into English beforehand, then
treat the task as monolingual retrieval (previously referred to as {\tt 1MT}). At recent 
evaluation campaigns like CLEF and NTCIR,\footnote{Most recent MLQA tracks were in 2008 (CLEF) and 2010 (NTCIR).} 
almost all teams simply obtained the one-best question translation, treating some online MT system as a black 
box~\cite{Adafre:2009aa,Hartrumpf:2009aa,Martinez-Gonzalez:2009aa,Lin:2010aa,Shima:2010aa}, with few notable 
exceptions that took term importance~\cite{Ren:aa}, or semantics~\cite{Munoz-Terol:2009aa} into account. Even for
non-factoid MLQA, most prior work does not focus on the translation component~\cite{Luo:2013aa,Chaturvedi:2014aa}.


\smallskip \noindent \textbf{Contributions.}
Ture and Lin described three methods for translating queries into the collection language 
in a probabilistic manner, improving \emph{document retrieval} effectiveness over a one-best translation 
approach~\shortcite{Ture:2014aa}.
Extending this idea to MLQA appears as a logical next step, yet most prior work
relies solely on the one-best translation of questions or answers~\cite{Ko:2010ab,Garcia-Cumbreras:2012aa,Chaturvedi:2014aa},
or selects the best translation out of few options~\cite{Sacaleanu:2008aa,Mitamura:2006aa}.
Mehdad et al. reported improvements by including the top ten translations (instead of the single best) and 
computing a distance-based entailment score with each~\shortcite{Mehdad:2010aa}. While Espla-Gomis et al. argue 
that using MT as a black box is more convenient (and modular)~\shortcite{Espla-Gomis:2012aa}, there are potential 
benefits from a closer integration between statistical MT and multilingual retrieval~\cite{Nie:2010}. To the best of 
our knowledge, there is no prior work in the literature, where the \emph{\textbf{optimal query and/or answer translation is 
learned via machine learning}}. This is our main contribution, with which we outperform the state of the art.

In addition to learning the optimal translation, we \emph{\textbf{learn the optimal subset of the training data for a given 
task}}, where the criteria of whether we include a certain data instance is based on either the source 
language of the sentence, or the language in which the sentence was annotated. Training data selection strategies 
have not been studied extensively in the QA literature, therefore the effectiveness of our simple language-related 
criteria can provide useful insights to the community.

When there are multiple independent approaches for ranking question-answer pairs, it is required to perform a {\em 
post-retrieval merge}:\ each approach generates a ranked list of answers, which are then merged into a 
final ranked list. This type of system combination approach has been applied to various settings in QA research. 
Merging at the document-level is common in IR systems~(e.g., \cite{Tsai:2008aa}), and has shown to improve 
multilingual QA performance as well~\cite{Garcia-Cumbreras:2012aa}. Many QA systems combine answers obtained by different 
variants of the underlying model~(e.g., \cite{Brill:2001aa} for monolingual, \cite{Ko:2010aa,Ko:2010ab} for multilingual 
QA). We are not aware, however, of any prior work that has explored the \emph{\textbf{merging of answers that are 
generated by language-specific ranking models}}. Although this does not show increased effectiveness in
our experiments, we believe that it brings a new perspective to the problem.

\section{Approach}\label{sec:main}


Our work is focused on a specific stage of the QA pipeline, namely \emph{answer ranking}:\
Given a natural-language question $q$ and $k$ candidate answers $s_1, \ldots, s_k$, we score each answer in terms of its relevance 
to $q$. In our case, candidate answers are sentences extracted from all documents retrieved in the previous stage of the 
pipeline (using Indri~\cite{Metzler:2005}). Hereafter, sentence and answer might be used interchangeably.

While our approach is not language-specific, we assume (for simplicity) that questions are in English, whereas sentences are 
in either English, Arabic, or Chinese. Non-English answers are translated back into English before 
returning to user. 

Our approach is not limited to any question type, factoid or non-factoid. Our main motivation is to provide good QA quality on
any multilingual Web collection. This entails finding answers to questions where there is no single answer, and for which human 
agreement is low. We aim to build a system that can successfully retrieve relevant information from open-domain and 
informal-language content. In this scenario, two assumptions made by many of the prior approaches fail:

1) We can accurately classify questions via template patterns (Chaturvedi et al. argue that this does not hold for non-factoid
questions~\shortcite{Chaturvedi:2014aa})\\
\indent 2) We can accurately determine the relevance of an answer, based on its automatic translation into English (Wees et al. 
show how recall decreases when translating user-generated text~\shortcite{Wees:2015aa})

To avoid these assumptions, we opted for a more adaptable approach, in which question-answer relevance is modeled 
as a function of features, intended to capture the relationship between the question 
and sentence text. Also, instead of relying solely on a single potentially incorrect English translation, we increase our chances of 
a hit by translating both the question and the candidate answer, using four different translation methods.
Our main features, described throughout this section, are based on lexical similarity computed using these translations. 
The classifier is trained on a large number of question-answer pairs, each labeled by a human annotator with a 
binary relevance label.\footnote{Annotators score each answer from 1 to 5. We label any score of 3 or higher as relevant.} 
\ignore{We also show that restricting the classifier to a specific subset of training data can result in a much better model of relevance.}



%


\subsection{Representation}\label{sec:repr}

In MLQA, since questions and answers are in different languages, most approaches translate both 
into an intermediary language (usually English). Due to the error-prone nature of MT, 
valuable information often gets ``lost in translation''. 
These errors are especially noticeable when translating informal
text or less-studied languages~\cite{Wees:2015aa}.

\smallskip \noindent \textbf{Translation Direction.}
We perform a \emph{two-way translation} to better retain the original meaning:
in addition to translating each non-English sentence into English, 
we also translate the English questions into Arabic and Chinese (using multiple translation methods, described 
below). For each question-answer pair, we have two ``views'':\ comparing translated question to the 
original sentence (i.e., {\em collection-language} (CL) view); and comparing original question to the 
translated sentence (i.e., {\em question-language} (QL) view). 


\smallskip \noindent \textbf{Translation Method.}
When translating text for retrieval tasks like QA, including a variety of alternative translations is as important as finding the most accurate 
translation, especially for non-factoid questions, where capturing (potentially multiple) underlying topics is essential.
Recent work in cross-language IR (CLIR) has shown that 
incorporating probabilities from the internal representations of an MT system to ``translate'' the question can accomplish this, 
outperforming standard one-best translation~\cite{Ture:2014aa}. 
We hypothesize that these improvements transfer to multilingual QA as well. 

In addition to \emph{translation directions}, we explored four \emph{translation methods} for converting 
the English question into a probabilistic representation (in Arabic and Chinese). Each method builds a probability distribution for 
every question word, expressing the translation space in the collection language.
More details of first three methods can be found in \cite{Ture:2014aa}, while fourth method is a novel query translation
method adapted from the neural network translation model described in \cite{Devlin:2014}.

\noindent \underline{\emph{Word}}: In MT, a word alignment is a many-to-many mapping between source- and target-language words,
learned without supervision, at the beginning of the training pipeline~\cite{Och:2003a}.
These alignments can be converted into word translation probabilities for CLIR~\cite{Darwish:2003aa}. \\
For example, in an English-Arabic parallel corpus, if an English word appears $m$ times in total and is aligned to a certain Arabic 
word $k$ times, we assign a probability of $\frac{k}{m}$ for this translation. This simple idea has performed greatly in IR for 
generating a probability distribution for query word translations.\\
\underline{\emph{Grammar}}: Probabilities are derived from a synchronous context-free grammar, which is a typical
translation model found in MT systems~\cite{Ture:2014aa}. 
The grammar contains rules $r$ that follow the form {\tt $\alpha$|$\beta$|$\mathcal{A}$|$\ell(r)$}, 
stating that source-language word $\alpha$ can be translated into target-language word $\beta$ with an associated likelihood value 
$\ell(r)$ ($\mathcal{A}$ represents word alignments).
For each rule $r$ that applies to the question, we identify each source word $s_j$. From the word
alignment information included in the rule, we can find all target words that $s_j$ is aligned to.
By processing all the rules to accumulate likelihood values, we construct translation probabilities for each word in the question.\\
\underline{\emph{10-best}}: Statistical MT systems retrieve a ranked list of translations, not a single best.
Ture and Lin exploited this to obtain word translation probabilities from the top 10 translations of the question~\shortcite{Ture:2014aa}.
For each question word $w$, we can extract which grammar rules were used to produce the translation -- once we have the
rules, word alignments allow us to find all target-language words that $w$ translates into.
By doing this for each question translation, we construct a probability distribution that defines the translation space of each
question word.\\
\underline{\emph{Context}}: Neural network-based MT models learn context-dependent word translation probabilities -- 
the probability of a target word is dependent on the source word it aligns to, as well as a 5-word window of context~\cite{Devlin:2014}.
For example, if the Spanish word ``placer'' is aligned to the English word ``pleasure'', the model will not only learn from this word-to-word 
alignment but also consider the source sentence context (e.g., ``Fue un placer conocerte y tenerte unos meses.''). However, since short
questions might lack full sentence context, our model should have the flexibility to translate under partial
or no context. Instead of training the NN-base translation model on full, well-formed sentences, we custom-fit it for question translation:\ 
words in the context window are randomly masked by replacing it with a special filler token {\tt <F>}. This teaches the model how to 
accurately translate with full, partial context, or no context. For the above example, we generate partial contexts such as ``fue un placer 
{\tt <F>} y'' or ``{\tt <F>} {\tt <F>} placer conocerte y''. Since there are many possibilities, if the context window is large, we randomly sample a few of 
the possibilities (e.g., 4 out of 9) per training word.

\begin{figure}
\centering
\includegraphics[scale=0.5]{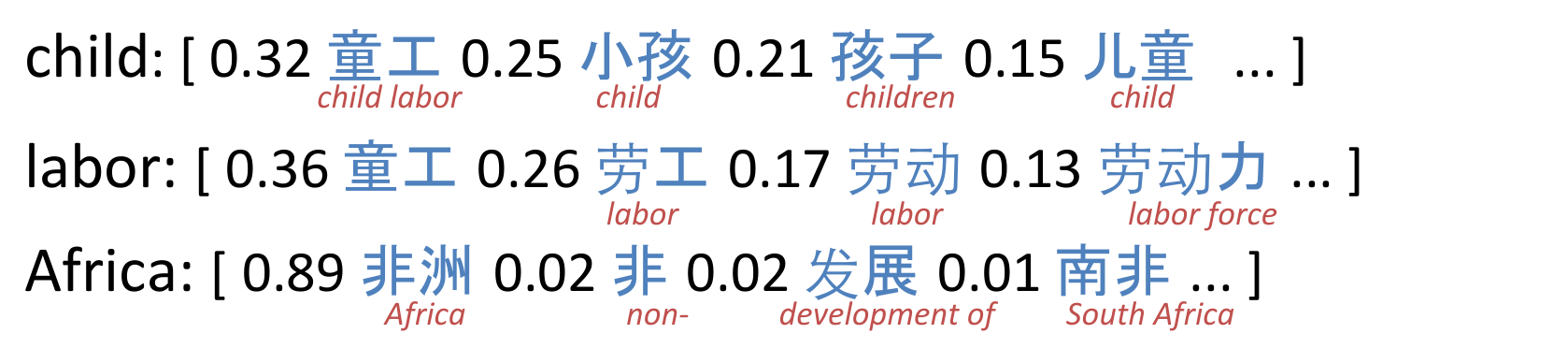}
\caption{Probabilistic grammar-based translation of example question. The example question ``Tell me about child labor in Africa" is simplified by our 
preprocessing engine to ``child labor africa".} 
 \label{fig:example}
\end{figure}

In Figure~\ref{fig:example}, we display the probabilistic structure produced the \emph{grammar-based} translation method, when implemented as 
described above. Each English word in the question is translated into a probabilistic structure, consisting of Chinese words and corresponding
probabilities that represent how much weight the method decides to put on that specific word. Similar structures are learned with the other three 
translation methods.

We are not aware of any other MLQA approach that represents the question-answer pair based on their probabilistic translation
space.
 
\subsection{Features}\label{sec:feat}


Given two different translation directions (\emph{CL} and \emph{QL}), and four different translation methods (\emph{Word}, 
\emph{Grammar}, \emph{10-best}, \emph{Context}), our strategy is to leverage a machine learning process to determine how 
helpful each signal is with respect to the end task. For this, we introduced separate question-answer similarity features based 
on each combination of translation direction and method. 


\smallskip \noindent {\bf Collection-language Features.} In order to compute a single real-valued vector to represent the question in the collection language 
(\emph{LexCL}), we start with the probabilistic structure representing the question translation (e.g., Figure~\ref{fig:example} is one such 
structure when the translation method is \emph{grammar-based}). For each word in the collection-language vocabulary, we compute a weight 
by averaging its probability across the terms in the probabilistic structure.
\begin{equation}\label{eq:v1}
v_{q_\textrm{grammar}}(w) = \textrm{avg}_{i} \textrm{Pr}(w|q_i)
\end{equation}
where $w$ is a non-Engish word and $\textrm{Pr}(w|q_i)$ is the probability of $w$ in the probability distribution corresponding to the 
$i^\textrm{th}$ query term.

\begin{figure}[h!]
\centering
\includegraphics[scale=0.45]{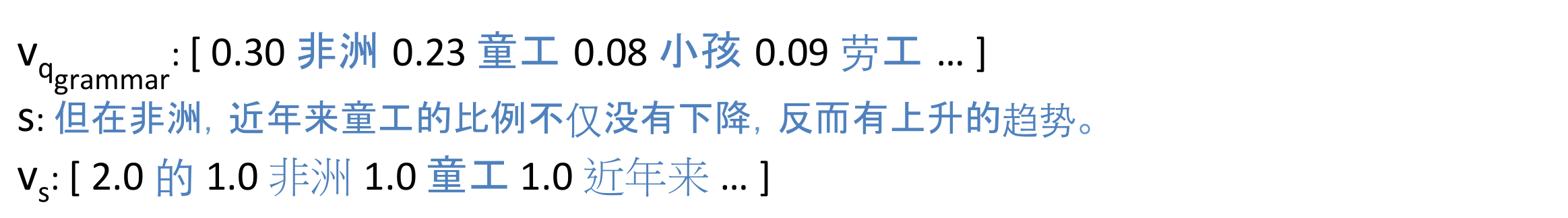}
\caption{Vector representation of grammar-translated question ($q_{\textrm{grammar}}$) and sentence ($s$).}
 \label{fig:vector}
\end{figure}

Figure~\ref{fig:vector} shows the real-valued vector computed based on the probabilistic question translation in Figure~\ref{fig:example}. 
The Chinese word translated as ``child labor'' has a weight of 0.32, 0.36, and 0 in the probability distributions of the three query terms, respectively. 
Averaging these three values results in the final weight of 0.23 in $v_{q_\textrm{grammar}}$ in Figure~\ref{fig:vector}. Notice that these
weights are normalized by construction.

Similarly, a candidate answer $s$ in Chinese is represented by normalized word frequencies:
\begin{equation}\label{eq:v2}
v_s(w) = \frac{\textrm{freq}(w|s)}{\sum_{w^\prime} \textrm{freq}(w^\prime|s)}
\end{equation}

Given the two vectors, we compute the cosine similarity.
Same process is repeated for the 
other three translation methods. The four lexical collection-language similarity features are collectively called \emph{LexCL}.

\smallskip \noindent {\bf Question-language Features.}
As mentioned before, we also obtain a similarity value by translating the sentence ($s_{\textrm{1best}}$) 
and computing the cosine similarity with the original question ($q$). $v_q$ and $v_{s_{\textrm{1best}}}$ are computed
using Equation~\ref{eq:v2}. Although it is possible to translate the 
sentence into English using the same four methods, we only used the one-best translation due to the computational 
cost. Hence, we have only one lexical similarity feature in the QL view (call \emph{LexQL}). 

The computation process for the five \emph{lexical similarity} features is summarized in Table~\ref{tab:features}.
After computation, feature weights are learned via a maximum-entropy model.\footnote{Support vector machines 
yielded worse results.} Although not included in the figure or table, we also include the 
same set of features from the sentence preceding the answer (within the corresponding forum post), in order to represent the larger discourse.

\begin{table}
\begin{tabular}{|c|c|c|c|}
\hline
\textbf{{\small Feature}} 	& \textbf{{\small Question}} & \textbf{{\small Sentence}} & \textbf{{\small Feature}}\\
\textbf{{\small category}} 	& \textbf{{\small repr. }($v_{q'}$)} 	  & \textbf{{\small repr.} ($v_{s'}$)} 	    & \textbf{{\small Value}}\\ \hline
\multirow{4}{*}{\emph{LexCL}} & $v_{q_\textrm{word}}$ & $v_s$ &\\\cline{2-3}
				  & $v_{q_\textrm{10best}}$ & $v_s$ &  \begin{tabular}[c]{@{}c@{}}cosine\\ $(v_{q'},v_{s'})$\end{tabular} \\ \cline{2-3}
				  & $v_{q_\textrm{context}}$ & $v_s$ &  \\ \cline{2-3}
				  & $v_{q_\textrm{grammar}}$ & $v_s$ & \\ \cline{1-3}
\emph{LexQL} 				& $v_q$ & $v_{s_\textrm{1best}}$ & \\ \hline
\end{tabular}
  \caption{List of features used in {\tt L2T}, and how the values are computed from vector representations.}
  \label{tab:features}
\end{table}
    \vspace{-0.2cm}

\subsection{Data Selection} \label{sec:data}
In order to train a machine learning model with our novel features, we need positive and negative examples 
of question-answer pairs (i.e., ($q,s$)). For this, for each training question, our approach is to hire human annotators to label 
sentences retrieved from the non-English collections used in our evaluation. It is possible to label
the sentences in the source language (i.e., Arabic or Chinese) or in the question language (i.e., translated into English).
In this section, we explore the question of whether it is useful to distinguish between these two independently created
labels, and whether this redundancy can be used to improve the machine learning process. 

We hypothesize two reasons why selecting training data based on language might benefit MLQA:\\
i) The translation of non-English candidate answers might lack in quality, so 
annotators are likely to judge some relevant answers as non-relevant. Hence, training a classifier on this data might lead to a 
tendency to favor English answers.\\
ii) For the question-answer pairs that were annotated in both languages, we can remove noisy (i.e., labeled inconsistently by 
annotators) instances from the training set.

The question of annotation is an unavoidable part of evaluation of 
MLQA systems, so finding the optimal subset for training is a relevant problem.
In order to explore further, we generated six subsets with respect to (a) the 
original \emph{language} of the answer, or (b) the language of \emph{annotation} (i.e., based on original text or its
English translation):\\
\underline{\em en}: Sentences from the English corpus.\\
\underline{\em  ar/ch}: Sentences from the Arabic / Chinese corpus (regardless of how it was judged).\\
\underline{\em consist}: All sentences except those that were judged inconsistently.\\
\underline{\em src+}: Sentences judged only in original text, or judged in both consistently.\\
\underline{\em en+}: Sentences that are either judged only in English, or judged in both original and English translation consistently.\\
\underline{\em all}: All sentences.

These subsets were determined based on linguistically motivated heuristics, but choosing the most suitable one (for a given task)
is done via machine learning (see Section~\ref{sec:eval}). 


\subsection{Language-specific Ranking} \label{sec:oneclassperlang}

Scoring Arabic sentences with respect to a question is inherently different than scoring English (or Chinese) sentences. The  
quality of resources, grammar, etc., as well as other internal dynamics might differ greatly across languages. We hypothesize that there 
is no one-size-fits-all model, so the parameters that work best for English retrieval might not be as useful when scoring 
sentences in Arabic, and/or Chinese.

Our proposed solution is to apply a separate classifier, custom-tuned to each collection, and retrieve three \emph{single-language ranked lists} 
(i.e., in English, Arabic, and Chinese).
In addition to comparing each custom-tuned, language-specific classifier to a single, language-independent one, we also use this idea 
to propose an approach for MLQA:\\
\underline{{\tt L2CT}}$(n)$ Retrieve answers from each language using separate classifiers (call these lists English-only, Arabic-only,
and Chinese-only), take the best answers from each language, then merge them into a mixed-language set of $n$ answers.\\
\indent We compare this to the standard approach:\\
\underline{{\tt L2T}}$(n)$ Retrieve up to $n$ mixed-language answers using a single classifier.

Four heuristics were explored for merging lists in the {\tt L2CT} approach.\footnote{In addition to these heuristics, 
the optimal merge could be learned from training data, as a ``learning to rank'' problem. This is out of the scope of this paper, but we plan to explore 
the idea in the future.} 
Two common approaches are \emph{uniform} and \emph{alternate} merging~\cite{Savoy:2004aa}:\\
\underline{{\em Uniform}}: A straightforward merge can be achieved by using the classifier scores (i.e., probability of answer relevance, given question)
to sort all answers, across all languages, and include the top $n$ in the final list of answers. Classifier scores are normalized into the [0,1] range
for comparability.\\
\underline{{\em Alternate}}: We alternate between the lists, picking one answer at a time from each, stopping when the limit $n$ has been reached.

Since answers are expected in English, there is a natural preference for answers that were originally written English, avoiding 
noisy text due to translation errors. However, it is also important not to restrict answers entirely to English sources, since that would defeat 
the purpose of searching in a multilingual collection. We implemented the following methods to account for language preferences:\\
\underline{{\em English first}}: We keep all sufficiently-confident (i.e., normalized score above a fixed threshold) answers from the 
English-only list first, and start including answers from Arabic- and Chinese-only lists only if the limit of $n$ answers has not been reached. \\
\underline{{\em Weighted}}: Similar to {\em Uniform}, but we weight the normalized scores before sorting. 
The optimal weights can be learned by using a grid-search procedure and a cross-validation split.


\section{Evaluation}\label{sec:eval}
\ignore{In this section, we describe the evaluation of our multilingual QA approach.}
In order to perform controlled experiments and gain more 
insights, we split our evaluation into four separate tasks:\ three tasks focus on retrieving answers from posts written in a specified language 
(\emph{English-only}, \emph{Arabic-only}, or \emph{Chinese-only})~\footnote{Shortened as \emph{Eng}, \emph{Arz}, and \emph{Cmn}, 
respectively.}, and the last task is not restricted to any language (\emph{Mixed-language}). 
All experiments were conducted on the DARPA BOLT-IR task. 
The collection consists of 12.6M Arabic, 7.5M Chinese, and 9.6M English Web forum posts. All 
runs use a set of 45 non-factoid (mostly opinion and causal) English questions, from a range of topics.
All questions and forum posts were processed with an information extraction (IE) toolkit~\cite{Boschee:2005aa}, 
which performs sentence-splitting, named entity recognition, coreference resolution, parsing, and part-of-speech tagging. 

All non-English posts were translated into English (one-best only), and all questions were translated into Arabic and Chinese 
(probabilistic translation methods from Section~\ref{sec:repr}).
For all experiments, we used the same state-of-the-art English$\leftrightarrow$Arabic (En-Ar) and English$\leftrightarrow$Chinese 
(En-Ch) MT systems~\cite{Devlin:2014}. 
Models were trained on parallel corpora from NIST OpenMT 2012, in addition to parallel forum data collected 
as part of the BOLT program (10M En-Ar words; 30M En-Ch words). Word alignments were learned with 
GIZA++~\cite{Och:2003} (five iterations of IBM Models 1--4 and HMM). 

After all preprocessing, features were computed using the original post and question text, and their translations.
Training data were created by having annotators label all sentences of the top 200 documents retrieved by Indri from 
each collection (for each question). 
Due to the nature of retrieval tasks, training data usually contains an unbalanced portion of negative examples. 
Hence, we split the data into balanced subsets (each sharing the same set of positively labeled data) 
and train multiple classifiers, then take a majority vote when predicting.

For testing, we froze the set of candidate answers and applied the trained classifier to each question-answer pair, generating
a ranked list of answers for each question. This ranked list was evaluated by average precision (AP).\footnote{Many other metrics
(e.g., NDCG, R-precision) were explored during BOLT, and results were very similar.} 
Due to the size and redundancy of the collections, we sometimes end 
up with over 1000 known relevant answers for a question. So it is neither reasonable nor meaningful to compute AP until we
reach 100\% recall (e.g., 11-point AP) for these cases. Instead, we computed AP-$k$, by accumulating precision values 
at every relevant answer until we get $k$ relevant answers.\footnote{$k$ was fixed to 20 in our evaluation, although we
verified that conclusions do not change with varying $k$.} In order to provide a single metric for the test set, it is common
to report the mean average precision (MAP), which in this case is the average of the AP-$k$ values across all questions.

\smallskip \noindent \textbf{Baseline.}
As described earlier, the baseline system computes similarity between question text and the one-best translation of the candidate 
answer (we run the sentence through our state-of-the-art MT system). After translation, we compute similarity via scoring 
the match between the parse of the question text and the parse of the candidate answer, using our finely-tuned IE 
toolkit~[reference removed for anonymization].
This results in three different similarity features:\ matching the tree node similarity, 
edge similarity, and full tree similarity. Feature weights are then learned by training this classifier discriminatively on the 
training data described above. This already performs competitively, outperforming the simpler baseline where we compute 
a single similarity score between question and translated text, and matching the performance of the system by 
Chaturvedi et al.~on the BOLT evaluation~\shortcite{Chaturvedi:2014aa}. Baseline MAP values are 
reported on the leftmost column of Table~\ref{tab:controlled}.

\smallskip \noindent \textbf{Data effect.}
In the baseline approach, we do not perform any data selection, and use all available data for training the
classifier. In order to test our hypothesis that selecting a linguistically-motivated subset of the training data
might help, we used 10-fold cross-validation to choose the optimal data set (among seven options described in 
Section~\ref{sec:data}). Results indicate that including English or Arabic sentences when training a classifier 
for Chinese-only QA is a bad idea, since effectiveness increases when restricted to Chinese sentences ({\tt lang=ch}).
On the other hand, for the remaining three tasks, the most effective training data set is {\tt annot=en+consist}.
These selections are consistent across all ten folds, and the difference is statistically significant for all but Arabic-only.
The second column in Table~\ref{tab:controlled} displays the MAP achieved when data selection is applied
before training the baseline model.


\smallskip \noindent \textbf{Feature effect.}
To measure the impact of our novel features, we trained classifiers using either \emph{LexCL}, \emph{LexQL}, 
or \emph{both} feature sets (Section~\ref{sec:feat}). In these experiments, the data is fixed to the
optimal subset found earlier. Results are summarized on right side of Table~\ref{tab:controlled}. Statistically
significants improvements over {\em Baseline}/{\em Baseline+Data selection} are indicated with single/double 
underlining.
 
For Arabic-only QA, adding \emph{LexQL} features yields greatest improvements over the baseline, 
while the same statement holds for \emph{LexCL} features for the Chinese-only task.
For the English-only and mixed-language tasks, the most significant increase in MAP is observed 
with all of our probabilistic bilingual features. For all but Arabic-only QA, the MAP is statistically significantly 
better ($p<0.05$) than the baseline; for Chinese-only and mixed-language tasks, it also outperforms baseline 
plus data selection ($p<0.05$).\footnote{Note that bilingual features are not expected to help on the 
English-only task, and the improvements come solely from data selection.}
All of this indicates the effectiveness of our probabilistic question translation, 
as well as our data selection strategy.

\begin{table}[htb]
\centering
\begin{tabular}{c|c|c|l}
\textbf{Task} & \textbf{Base} & \textbf{+Data} & \textbf{+Feats}\\ \hline
Cmn & 0.416 & \underline{0.425} (\emph{ch})   & \underline{\underline{ 0.451 }} (\emph{LexCL})  \\ \cline{1-4}
Arz & 0.421 & 0.423 (\emph{en+}) 				& 0.425 (\emph{LexQL})  \\ \cline{1-4}
Eng & 0.637 & \underline{0.657} (\emph{en+}) 		& \underline{ 0.660 } (\emph{all}) \\ \hline
Mixed & 0.665 & \underline{0.675} (\emph{en+}) 	& \underline{\underline{ 0.681} } (\emph{all})\\ \hline
\end{tabular}
\caption{{\tt L2T} evaluated using MAP with 10-fold cross-validation for each task.
A statistically significant increase over Baseline/Base+Data is shown by single/double underlining ($p<0.05$).}
\label{tab:controlled}
\end{table}



Understanding the contribution of each of the four \emph{LexCL} features is also important. To gain insight, we 
trained a classifier using all {\em LexCL} features (using the optimal data subset learned earlier for each task), and then 
incrementally removed one of the features, and tested on the same task. 
This controlled experiment revealed that the \emph{word} translation feature is most useful for Chinese-only 
QA (i.e., removing it produces largest drop in MAP, 0.6 points), whereas \emph{context} translation appears 
to be most useful (by a slighter margin) in Arabic-only QA. In the former case, the diversity provided by
word translation might be useful at increasing recall in retrieving Chinese answers. In retrieving Arabic
answers, using context to disambiguate the translation might be useful at increasing precision. This result
further emphasizes the importance of a customized translation approach for MLQA.

Furthermore, to test the effectiveness of the 
probabilistic translation approach (Section~\ref{sec:repr}), we replaced all {\em LexCL} features with a 
single lexical similarity feature computed from the one-best question translation. This resulted
in lower MAP:\ 0.427 to 0.423 for Arabic-only, and 0.451 to 0.425 for Chinese-only task ($p < 0.01$), 
supporting the hypothesis that \emph{probabilistic translation is more effective than the widely-used 
one-best translation}. In fact, almost all gains in Chinese-only QA seems to be coming from the
probabilistic translation.

For a robustness test, we let cross-validation select the best combination of 
(\emph{data}, \emph{feature}), mimicking a less controlled, real-world setting. In this case, the best 
MAP for the Arabic-, Chinese-, English-only, and Mixed-language tasks are 0.403, 0.448, 0.657, and 
0.679, respectively. In all but Arabic-only, these are statistically significantly better ($p < 0.05$) than 
not tuning the feature set or training data (i.e., Baseline). This result suggests that our approach
can be used for any MLQA task out of the box, and provide improvements.

\smallskip \noindent \textbf{Learning to Custom Translate ({\tt L2CT}).}
We took the ranked list of answers output by each language-specific model, and merged all of them into a ranked list of mixed-language answers.
For the \emph{weighted} heuristic, we tried three values for the weight.
In Table~\ref{tab:multiple}, we see that training separate classifiers for each subtask does not bring overall improvements to 
the end task. Amongst merging strategies, the most effective were \emph{weighted} (weights for each query 
learned by performing a grid-search on other queries) and \emph{English first} -- however, both are statistically indistinguishable 
from the single classifier baseline. In the latter case, the percentage of English answers is highest (88\%), which might not be desirable.
Depending on the application, the ratio of languages can be adjusted with an appropriate merging method.
For instance, \emph{alternate} and \emph{norm} heuristics tend to represent languages almost equally.

\vspace{-0.07cm}
\begin{table}[h]
\begin{tabular}{cc|c|c|c}
\multicolumn{3}{c|}{\textbf{Approach}} & \textbf{(En-Ch-Ar)} \% & \textbf{MAP} \\ \hline
\multicolumn{3}{l|}{{\tt L2T}} & 64-19-16 & 0.681 \\ \hline\cline{1-5} 
\multicolumn{1}{c}{\multirow{6}{*}{{\tt L2CT}}} & \multicolumn{2}{|c|}{Uniform}  & 24-35-41 & 0.548 \\ \cline{2-5} 
\multicolumn{1}{c}{}& \multicolumn{2}{|c|}{Alt.} & 32-34-34 & 0.574 \\ \cline{2-5} 
\multicolumn{1}{c}{}& \multicolumn{2}{|c|}{Eng. First} & 88-6-6 & 0.668 \\ \cline{2-5} 
\multicolumn{1}{c|}{} & \multicolumn{1}{c|}{\multirow{3}{*}{Weight}} & 2 & 37-30-34 & 0.599 \\ \cline{3-5} 
\multicolumn{1}{c|}{} & \multicolumn{1}{c|}{} 					& 5 & 51-24-25 & 0.654 \\ \cline{3-5} 
\multicolumn{1}{c|}{} & \multicolumn{1}{c|}{} 					& 10 & 61-20-19 &  0.669\\
\end{tabular}
\caption{{\tt L2T} vs. {\tt L2CT} for multilingual QA.}
\label{tab:multiple}
\vspace{-0.2cm}
\end{table}

Even though we get lower MAP in the overall task, Table~\ref{tab:controlled} suggests that it is worthwhile
customizing classifiers for each subtask (e.g., the Chinese responses in the ranked list of {\tt L2CT} are more 
relevant than {\em Single}.). The question of how to effectively combine the results into a mixed-language list,
however, remains an open question.

\section{Conclusions}\label{sec:concl}
We introduced {\tt L2T}, a novel approach for MLQA, inspired from recent success in CLIR research.
To our knowledge, this is the first use of probabilistic translation methods for this task, and the first attempt
at using machine learning to learn the optimal question translation. 
\ignore{ We described four 
different translation methods to represent English questions in Arabic and Chinese.
vocabulary spaces, each of which adapts 
various stages of modern statistical MT systems to translate questions more effectively.}

We also proposed {\tt L2CT}, which uses language-specific classifiers to treat the ranking of English, 
Arabic, and Chinese answers as three separate subtasks, by applying a separate classifier for each language.
While post-retrieval merging has been studied in the past, we have not come across any work that applies this idea 
specifically to create a language-aware ranking for MLQA.


Our experimental analysis shows the importance of data selection when dealing with annotations on source and
translated text, and the effect of combining translation methods. 
{\tt L2T} improved answer ranking effectiveness significantly for Chinese-only, English-only, and mixed-language QA. 

Although results did not support the hypothesis that \emph{learning a custom classifier} for the retrieval of each language 
would outperform the \emph{single classifier} baseline, we think that more research is needed to fully 
understand how language-specific modeling can benefit MLQA. 
More sophisticated merging of multiple ranked lists of answers need to be explored. Learning to rank between answers 
from different languages might be more effective than 
heuristics. This would allow us to predict the final language ratio, based on many features (e.g., general
collection statistics, quality of candidate answers, question category and complexity, MT system confidence levels) 
to merge question-answer pairs.

An even more comprehensive use of machine learning would be to learn word-level translation scores, instead
of relying on translation probabilities from the bilingual dictionary, resulting in a fully customized translation. 
Similar approaches have appeared in learning-to-rank literature for monolingual IR~\cite{Bendersky:2010aa}, 
but not for multilingual retrieval. 
Another extension of this work would be to apply the same translation for translating answers into
the question language (in addition to question translation). By doing this, we would be able to capture the 
semantics of each answer much better, since we have discussed that one-best translation discards a lot of 
potentially useful information. 
\ignore{For this extension, the processing pipeline would need to handle the increased 
computational complexity.}

Finally, since one of the take-away messages of our work is that a deeper understanding of linguistic context can 
improve QA effectiveness via more sophisticated question translation, we are hoping to see even more improvements
by creating features based on word embeddings. One potential next step is to learn bilingual embeddings directly for 
the task of QA, for which we have started adapting some related work~\cite{Bai:2010aa}.

\section*{Acknowledgements}
Jacob Devlin has provided great help in the design and implementation of the context-based question translation 
approach. We would also like to thank the anonymous reviewers for their helpful feedback. 

%
\bibliographystyle{emnlp2016}
\bibliography{qa}

\end{document}